%% file: EUSAR-2026-Submission-v2.tex
\documentclass[10pt,english,oneside,twocolumn,a4paper]{article}
\usepackage{graphicx}
\usepackage[utf8]{inputenc}
\usepackage{times}
\fontfamily{ptm}\selectfont
\usepackage{tabularx}
\usepackage{ragged2e}
\usepackage[singlelinecheck=false]{caption}
\usepackage[T1]{fontenc}
\usepackage[numbers]{natbib}
\usepackage{amsmath}
\usepackage{amssymb}
\usepackage{graphicx}
\usepackage{babel}
\usepackage{tabularx} 
\usepackage{multirow}
\usepackage{subcaption} 
\usepackage{xcolor} 
\usepackage{enumitem}  
\usepackage{graphicx}
\usepackage[normalem]{ulem}
\useunder{\uline}{\ul}{}
\usepackage{algorithm}
\usepackage{algpseudocode}
\usepackage{hyperref}
\usepackage{lipsum}
\usepackage{float}

\newcommand\blfootnote[1]{%
  \begingroup
  \renewcommand\thefootnote{}\footnote{#1}%
  \addtocounter{footnote}{-1}%
  \endgroup
}
    
\RequirePackage[blocks]{authblk}
\makeatletter
\let\ps@plain\ps@empty
\def\@xivpt{14pt}
\def\@sect#1#2#3#4#5#6[#7]#8{%
  \ifnum #2<2
    \null\par\vskip-15pt
  \fi
  \ifnum #2>\c@secnumdepth 
    \let\@svsec\@empty
  \else
    \refstepcounter{#1}%
    \protected@edef\@svsec{%
      \ifnum #2<4
        \hb@xt@10mm{\csname the#1\endcsname}\relax
      \else
        \hb@xt@12mm{\csname the#1\endcsname}\relax
      \fi}%
  \fi
  \@tempskipa #5\relax
  \ifdim \@tempskipa>\z@
    \begingroup
      #6{%
        \@hangfrom{\hskip #3\relax\@svsec}%
          \interlinepenalty \@M #8\@@par}%
    \endgroup
    \csname #1mark\endcsname{#7}%
    \addcontentsline{toc}{#1}{%
      \ifnum #2>\c@secnumdepth \else  
        \protect\numberline{\csname the#1\endcsname}%
      \fi 
      #7}%
  \else
    \def\@svsechd{%
      #6{\hskip #3\relax
      \@svsec #8}%
      \csname #1mark\endcsname{#7}%
      \addcontentsline{toc}{#1}{%
        \ifnum #2>\c@secnumdepth \else
          \protect\numberline{\csname the#1\endcsname}%
        \fi
        #7}}%
  \fi
  \@xsect{#5}}
\renewcommand\LARGE{\@setfontsize\LARGE{16}{20}}
\def\abstract#1{\def\@abstract{#1}}
\def\abstractEn#1{\def\@abstractEn{#1}}
\def\titleEn#1{\def\@titleEn{#1}}
\def\@maketitle{%
  \newpage
  \null
  \let \footnote \thanks
    {\LARGE\bfseries\RaggedRight \@title \par}%
    \vskip 1\baselineskip%
    {\normalsize
      \@author\par}%
    \vskip 2\baselineskip%
    \vskip \baselineskip%
    {\section*{Abstract}
      \@abstract}%
  \par
  \vskip 3\baselineskip}

\renewcommand\section{\@startsection {section}{1}{\z@}%
                                   {-3.5ex \@plus -1ex \@minus -.2ex}%
                                   {\baselineskip}%
                                   {\normalfont\Large\bfseries\RaggedRight}}
\renewcommand\subsection{\@startsection{subsection}{2}{\z@}%
                                     {\baselineskip}%
                                     {1ex}%
                                     {\normalfont\large\bfseries\RaggedRight}}
\renewcommand\subsubsection{\@startsection{subsubsection}{3}{\z@}%
                                     {1\baselineskip}%
                                     {3bp}%
                                     {\normalfont\normalsize\bfseries\RaggedRight}}
\renewcommand\paragraph{\@startsection{paragraph}{4}{\z@}%
                                    {1\baselineskip\@plus1ex \@minus.2ex}%
                                    {3bp}%
                                    {\normalfont\normalsize\RaggedRight}}
\renewcommand\subparagraph{\@startsection{subparagraph}{5}{\parindent}%
                                       {3.25ex \@plus1ex \@minus .2ex}%
                                       {-1em}%
                                      {\normalfont\normalsize\bfseries\RaggedRight}}
\parindent\p@
\parskip\z@
\DeclareCaptionLabelSeparator{enskip}{\enskip}
\makeatother

\title{ZeroFlood: Flood Hazard Mapping from Single-Modality SAR Using Geo-Foundation Models}

\author[a]{Hyeongkyun Kim}
\author[b,c]{Orestis Oikonomou}
\affil[a]{German Aerospace Center (DLR), Remote Sensing Technology Institute, Oberpfaffenhofen, Germany}
\affil[b]{ETH Zurich, Seminar for Applied Mathematics, Zurich, Switzerland}
\affil[c]{ETH AI Center, Zürich, Switzerland}

\abstract{
Flood hazard mapping is essential for disaster prevention but remains challenging in data-scarce regions, where traditional hydrodynamic models require extensive geophysical inputs. This paper introduces \textit{ZeroFlood}, a framework that leverages Geo-Foundation Models (GeoFMs) to predict flood hazard maps using single-modality Earth Observation (EO) data, specifically SAR imagery. We construct a dataset that pairs EO data with flood hazard simulations across the European continent. Using this dataset, we evaluate several recent GeoFMs for the flood hazard segmentation task. Experimental results show that the best-performing model, TerraMind, achieves an F1-score of 88.36\%, outperforming supervised learning baselines by more than 3 percentage points. We shows the performance can be further improved by applying the Thinking-in-Modality (TiM) mechanism. These results demonstrate the potential of Geo-Foundation Models for data-driven flood hazard mapping using limited observational inputs. The dataset and experiment code are publicly available at \url{https://github.com/khyeongkyun/zeroflood}.
}
 
\begin{document}

\maketitle

\section{Introduction}

Floods are among the most frequent and devastating natural disasters worldwide. They cause significant loss of life, economic damage, and long-term social disruption \cite{A19, RS20, TSK21}. Consequently, there is an increasing need for pre-event mitigation strategies, commonly referred to as flood hazard mapping, which go beyond post-event flood response.

Traditional approaches to flood hazard mapping rely on hydrological and hydrodynamic simulations that require multiple heterogeneous data sources. For example, digital elevation models, precipitation records, land cover data, and river network information are commonly used \cite{DAB21, SSB15}. However, these approaches are difficult to apply in data-scarce regions, where flood impacts are often more severe due to the lack of reliable data.

\begin{figure*}[t!]
  \centering
  \includegraphics[width=0.98\textwidth]{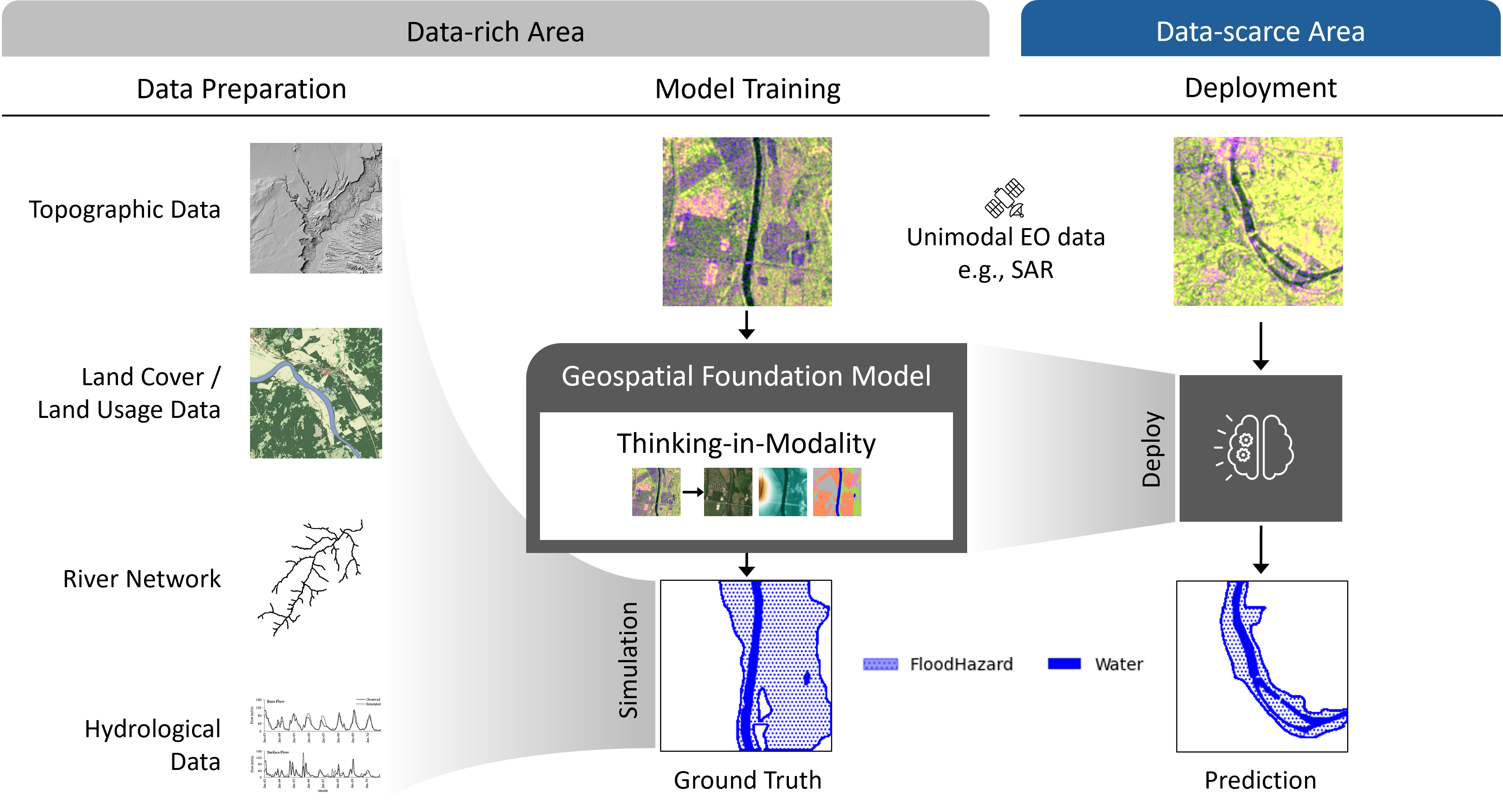}
  \caption{Overview of ZeroFlood framework. Data-rich regions provide the training dataset, which consists of flood hazard simulations and EO data (left). The GeoFM is trained to predict flood hazards using a single modality. TiM process supplements missing modalities during training (middle). The trained model can be deployed in data-scarce regions to predict flood hazard maps using only unimodal EO data (right).}
  \label{fig:zeroflood}
\end{figure*}

To address this limitation, we introduce \textit{ZeroFlood}, a framework that leverages a Geo-Foundation model (GeoFM) to predict flood hazards using unimodal Earth observation (EO) data, such as SAR imagery. As illustrated in \textbf{Figure~\ref{fig:zeroflood}}, we first align the EO datasets with the corresponding flood hazard simulation to build a training data set. This data set is then used to train six different neural networks and evaluate the flood hazard segmentation task.

Even under the single-modality setting, our results demonstrate the benefits of the Thinking-in-Modality (TiM) method, which compensates for missing modalities during inference by internally generating intermediate feature representations. These findings highlight the potential of deep learning approaches for the flood hazard prediction.

\vspace{0.5em}

This work addresses data scarcity in flood hazard mapping by combining EO data with geospatial foundation models.
Our contributions are summarized as follows:

\begin{itemize}
    \item We present ZeroFlood, a new dataset that pairs multimodal EO  data with flood hazard simulations, enabling data-driven research on flood hazard prediction.
    \item We demonstrate that flood hazard maps can be predicted directly from a single EO modality, specifically SAR imagery, offering a practical solution for flood risk assessment in data-scarce regions.
    \item We benchmark several recent GeoFMs on the flood hazard segmentation task and demonstrate their advantages over conventional supervised models.
    \item We analyze the impact of the TiM mechanism for compensating missing modalities and show that it further improves prediction performance in unimodal settings.
\end{itemize}

\blfootnote{This research extends the authors’ project organized by the European Space Agency (ESA) and IBM under the BlueSky Challenge and supported by the FAST-EO project, contract no.\ 4000143501/23/I-DT.}

\section{Related Work}


\textbf{Simulation Model for Flood Hazard Mapping.} Hydrological and hydrodynamic models play an important role in identifying flood-prone areas through physics-based simulations \cite{SSB15}. Recent research has introduced several models, such as Delft3D, InfoWorks ICM, LISFLOOD-FP, and CaMa-Flood, that simulate water flow and inundation using high-resolution digital elevation models, precipitation, land use, and river network data \cite{BIP20, DAB21, SJM21}. While these models provide accurate and physically interpretable results, they require extensive, high-quality input data and significant computational resources, which can limit their application in data-scarce or rapidly changing regions \cite{ASC20, NC20}. Studies highlight that the performance of simulation models is inherently constrained by the quantity and quality of input data, and large-scale implementations are often hindered by the trade-off between computational speed and accuracy \cite{ZSH16}. This dependency motivates learning-based approaches that infer flood hazard directly from remote-sensing observations, reducing reliance on dense physical inputs.
\vspace{0.5em}

\textbf{Machine Learning for Flood Hazard Mapping.} Unlike physics-based simulation models, Machine Learning (ML) approaches can learn complex and nonlinear relationships from diverse data sources, reducing the dependency on explicit physical modeling. Traditional ML methods, such as Support Vector Machines, Random Forests, and Ensemble models, have demonstrated high predictive accuracy \cite{MRM23, PLD22, SKH23}. More recently, deep learning methods, including MLP, CNN, and RNN architectures, have advanced the extraction of spatio-temporal features and improved model robustness and scalability \cite{DYF21, HG21, LXC16}. Nonetheless, their performance still relies on the availability of sufficient and diverse training data, as they are typically trained for a specific downstream task. Therefore, data dependency remains a key challenge for deployment in data-scarce regions \cite{FEA24}.
\vspace{0.5em}

\textbf{Geo-Foundation Models (GeoFMs).} GeoFMs are large-scale pre-trained models to unify EO modalities and enable downstream adaptations. They are built on diverse geospatial datasets, including optical, radar, and multispectral imagery, \cite{MJB25, WBX23}. To overcome label scarcity, GeoFMs typically leverage self-supervised learning techniques such as Masked Autoencoders, Contrastive Learning, and Self-distillation \cite{WBX23, FMG23, XWZ25, JYB25}. Owing to their generalized geospatial representations, GeoFMs demonstrate strong performance across various downstream tasks under low-data regimes, making them ideal for flood management applications in data-scarce environments \cite{DFC24, LLW23}. While multimodal and cross-modal approaches are becoming increasingly common for extensive feature representations \cite{JYB25, WBX23, XWZ25}, the practical applications of GeoFM for flood-related tasks remain post-event like flood detection. This motivates our exploration of using GeoFMs for pre-event of flood disasters.

\section{ZeroFlood Dataset}

\subsection{Data source \& processing}

The proposed model takes single-modal EO data as input and produces a flood hazard map as output. To construct the model input, we first curate data from the large-scale EO dataset \textit{TerraMesh} \cite{BFM25}. This dataset provides five EO modalities—Sentinel-1 RTC (S1RTC), Sentinel-2 Level 2A (S2L2A), Digital Elevation Model (DEM), Land Usage and Land Cover (LULC), and Normalized Difference Vegetation Index (NDVI)—with global coverage and precise spatial alignment across modalities. In our experiments, S1RTC data is used as the primary input modality. The remaining modalities, except NDVI, are used as auxiliary information for data quality assessment and result analysis.

For flood hazard masks, we use raster data simulated by \textit{LISFLOOD-FP} \cite{DAB21}. This dataset provides simulated flood hazard masks covering approximately 329,000 km of river networks across Europe and the Mediterranean basin, generated for nine different flood return periods. In this work, we use the 10-year return period masks as the ground truth for our experiments.

\begin{figure}[t!]
    \centering
    \includegraphics[width=\linewidth]{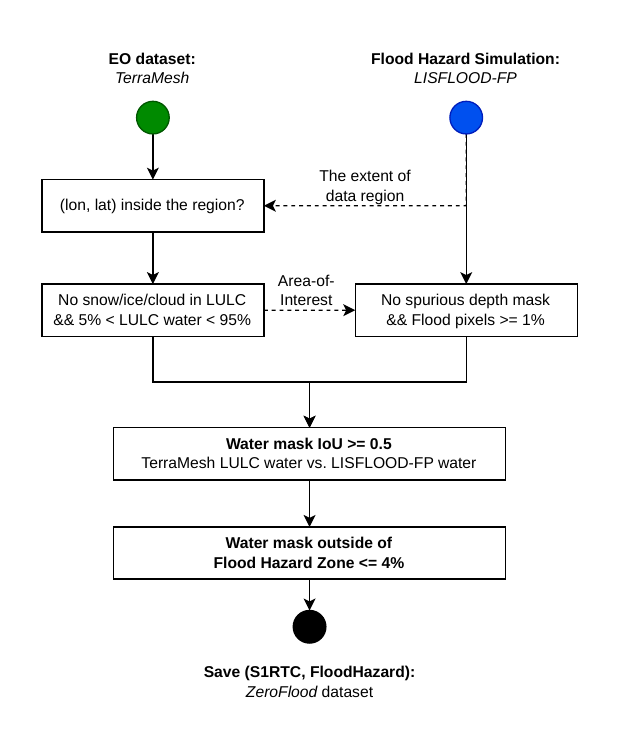}
    \caption{Data processing pipeline. \textit{TerraMesh} samples are filtered to ensure low noise and high information content (top left). \textit{LISFLOOD-FP} samples are cropped according to the area of interest (top right). The filtered samples are finally assessed and stored in the \textit{ZeroFlood} dataset (bottom).}
    \label{fig:data-processing}
\end{figure}

These two datasets differ in both regional coverage and spatial resolution. The \textit{TerraMesh} S1RTC data has a spatial resolution of 10 m and global coverage, whereas the \textit{LISFLOOD-FP} masks have a resolution of 3 arc-seconds ($\sim$90 m) and cover only the European region. To align their spatial characteristics and improve data quality, we apply several processing steps, as illustrated in \textbf{Figure~\ref{fig:data-processing}}.

First, samples from each dataset are filtered according to predefined rules. During this step, spatial information is exchanged between the datasets to synchronize the area of interest. Therefore, the samples are spatially aligned and cropped to the same regions. Finally, using the \textit{TerraMesh} LULC data together with the \textit{LISFLOOD-FP} water/flood masks, we select samples that clearly represent water bodies and flood-prone regions. Specifically, the first condition identifies the identical presence of water bodies, while the second condition removes samples where the flood hazard region is narrower than the corresponding LULC water body.

\subsection{Data characteristics}

\begin{figure}[t!]
    \centering
    \includegraphics[width=\linewidth]{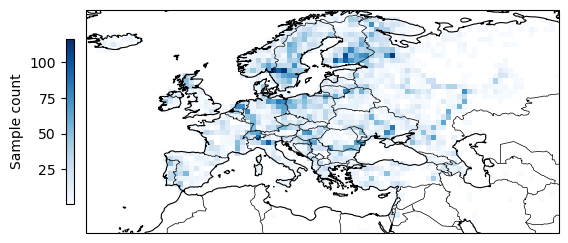}
    \caption{Spatial heat map of training samples in a one-degree resolution.}
    \label{fig:heatmap_train}
\end{figure}

Following the data processing procedure, the regional coverage of the \textit{ZeroFlood} dataset is limited to the European continent. \textbf{Figure~\ref{fig:heatmap_train}} shows the distribution of training samples across this region. The validation and test samples follow similar distribution characteristics.

In addition, the flood hazard masks obtained from the source data are not explicitly distinguishable from existing water bodies. As a result, some samples contain flood hazard masks that are nearly identical to the water bodies. We include these samples in the dataset to improve model robustness, as this allows the model to learn features of both flood-prone and non-flood-prone regions. As summarized in \textbf{Table~\ref{tab:dataset_stats}}, the presented \textit{ZeroFlood} dataset contains 20,868 samples for training, 5,217 samples for validation, and 226 samples for testing. The size of the image and the mask in all samples is $264\times264$ with a spatial resolution of 10 m. The total data volume is approximately 42~GB.

\input{tabs/dataset_stats}

\section{Experimental Setup}

\subsection{Baseline \& GeoFMs}

We evaluate the effectiveness of several GeoFMs by fine-tuning four representative models: SSL4EO-MAE \cite{WBX23}, CROMA \cite{FMG23}, DOFA \cite{XWZ25}, and TerraMind \cite{JYB25}. Their performance is compared with two widely used baseline architectures, U-Net \cite{RFB15} and Vision Transformer (ViT) \cite{DBK20}, which are trained in a supervised learning manner.

\begin{figure}[t!] \includegraphics[width=0.98\columnwidth]{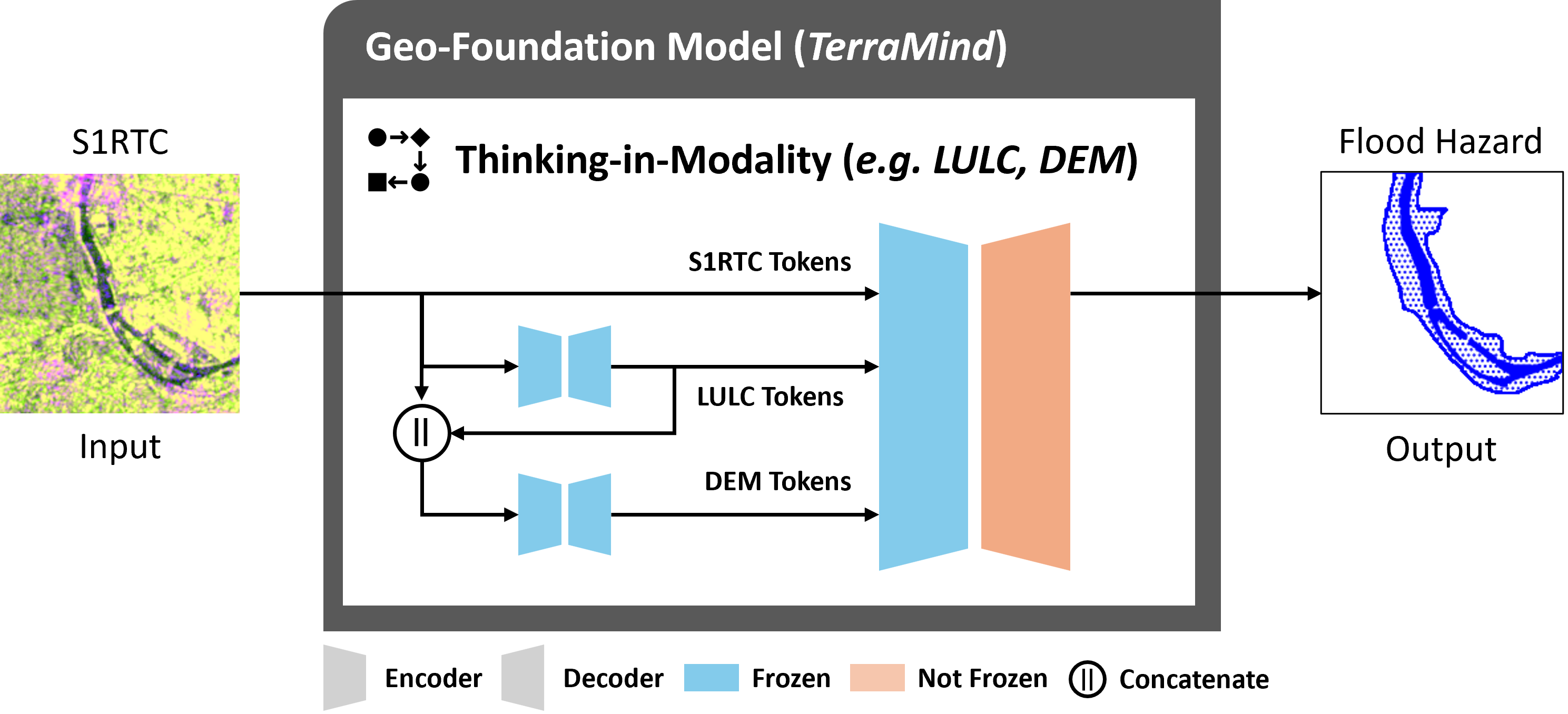}
  \caption{Forward process of Thinking-in-Modality. LULC and DEM tokens are sequentially generated from the S1RTC input.}
  \label{fig:diagram-tim}
\end{figure}

We further investigate the effectiveness of the Thinking-in-Modality (TiM) process to compensate for missing modalities and improve model reasoning capabilities. TiM was first introduced in TerraMind \cite{JYB25}, which is trained on multiple Earth Observation (EO) modalities and learns a shared token space with any-to-any generation. As illustrated in \textbf{Figure~\ref{fig:diagram-tim}}, the input modality S1RTC is first used to generate LULC tokens. Subsequently, through a chained generation process, both tokens are utilized to generate DEM tokens. All generated and original tokens are then jointly fed into the model to predict flood hazards. Based on this mechanism, we hypothesize that missing modalities can be approximated and used to improve flood hazard mapping. In this experiment, we incorporate three modalities—S2L2A, DEM, and LULC—within the TiM process.

\subsection{Evaluation Metrics}

To assess how accurately the model predicts flood hazard pixels, we adopt the following three evaluation metrics \cite{DAB21}. All scores are scaled to the range of 0 to 100, where higher values indicate better performance.
\vspace{0.5em}

\textbf{Hit Rate}, which corresponds to the \textit{Recall} metric in segmentation tasks, measures the proportion of ground-truth flood hazard pixels ($F_{gt}$) correctly predicted by the model ($F_{pred}$):

\[
HitRate = \frac{F_{gt} \cap F_{pred}}{F_{gt}}
\]

Since \textit{Hit Rate} alone does not penalize over-prediction, we additionally use \textbf{True Alarm}, which corresponds to \textit{Precision}. Measures the reliability of predicted flood pixels:

\[
TrueAlarm = \frac{F_{gt} \cap F_{pred}}{F_{pred}}
\]

To summarize overall performance, we compute \textbf{F1-score}, which balances \textit{Hit Rate} and \textit{True Alarm}:

\[
F1 = 2\times\frac{HitRate \times TrueAlarm}{HitRate + TrueAlarm}
\]

\subsection{Training Strategy}

For the baseline models, U-Net and ViT are trained from scratch. For the GeoFM-based models, the encoder is frozen during training and only the decoder weights are updated. Except for the U-Net baseline, all model encoders connect to UPerNet \cite{XLZ18} as a decoder. We train the model for 80 epochs, and the model with the highest mIoU in the validation set is selected for evaluation. When the input size of the model differs from the data resolution, sliding-window inference is applied as follows \cite{MJB25}. All experiments are conducted on a single NVIDIA A100 GPU node. Training takes less than one day, while inference on test samples requires less than one minute.

\section{Results}

We first evaluate flood hazard mapping performance using two supervised baselines, U-Net and ViT, together with four GeoFM-based models, as summarized in \textbf{Table~\ref{tab:main_metrics}}. For the TiM process, LULC are selected as an auxiliary modality. In general, most GeoFM-based models outperform supervised baselines. Among the evaluated models, TerraMind achieves an \textit{F1-score} of 88.36\%, outperforming the second best GeoFM model DOFA by 1.46 percentage points (pp). When the TiM process is activated, the performance improves further up to 89.12\%, yielding the best overall result.

A similar performance trend can be observed for both \textit{HitRate} and \textit{TrueAlarm}. Particularly, \textit{HitRate} is consistently lower than \textit{TrueAlarm} across all models. This pattern suggests that the models adopt a conservative prediction strategy, tending to avoid over-predicting flood hazards. The gap is largest for CROMA (8.23 pp) and smallest for TerraMind TiM (1.35 pp), indicating a more balanced prediction behavior for the latter. 

\input{tabs/main_metrics}

\begin{figure}[t!]
    \centering
    \includegraphics[width=\linewidth]{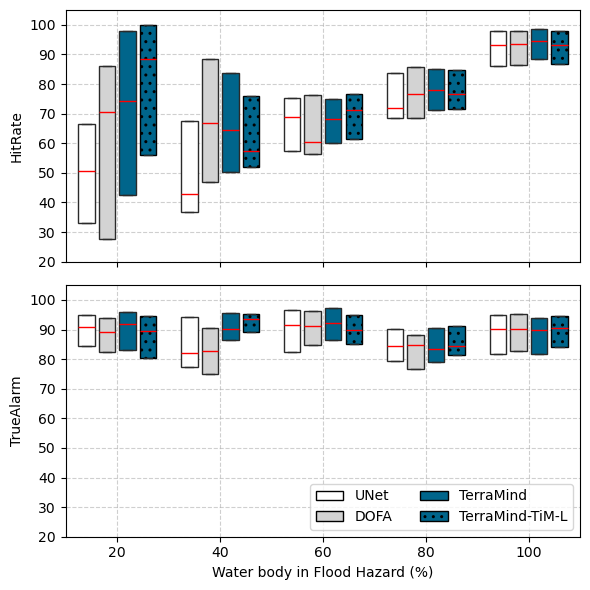}
    \caption{Distribution of model scores by the percentile of water bodies within flood hazard areas. The box indicates the upper and lower quartiles, and the red line inside the box represents the median.}
    \label{fig:perf_vs_waterbody}
\end{figure}

Furthermore, we analyze model performance with respect to the percentage of water bodies within flood hazard masks, as shown in \textbf{Figure~\ref{fig:perf_vs_waterbody}}. The plot confirms the trend, where \textit{TrueAlarm} consistently exceeds \textit{HitRate} across different water-body ratios. All models received higher scores as the proportion of water bodies within the flood risk area increased, indicating that the models have a strong tendency to predict flood-prone areas based on the shape of the water bodies. However, the \textit{HitRate} decreases noticeably as the water-body ratio becomes lower, whereas \textit{TrueAlarm} remains relatively stable across models. Especially, TerraMind TiM maintains more stable performance than the other models in scenarios with low water-body ratios. This improvement may be attributed to the TiM mechanism, where internally generated modalities can enhance terrain-related information and help the model better capture the relationship between surface characteristics and flood hazards.

\subsection{Ablation study on TiM}

\begin{figure*}[t!]
    \centering
    \begin{subfigure}[t]{0.32\linewidth}
        \centering
        \includegraphics[width=\linewidth]{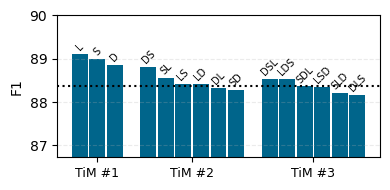}
    \end{subfigure}
    \hfill
    \begin{subfigure}[t]{0.32\linewidth}
        \centering
        \includegraphics[width=\linewidth]{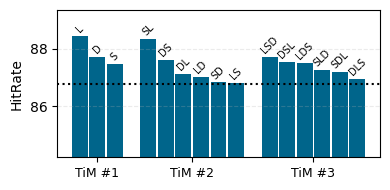}
    \end{subfigure}
    \hfill
    \begin{subfigure}[t]{0.32\linewidth}
        \centering
        \includegraphics[width=\linewidth]{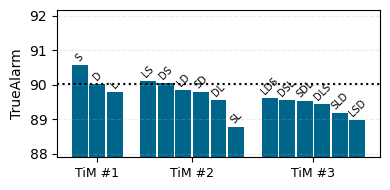}
    \end{subfigure}
    \caption{Performance variation with respect to the number and order of TiM. Generated modalities are represented by symbols: S (S2L2A), D (DEM), and L (LULC). The dotted horizontal line denotes the performance of the base TerraMind model without TiM.}
    \label{fig:tim}
\end{figure*}

We further analyze the impact of different TiM configurations, as illustrated in \textbf{Figure~\ref{fig:tim}}, where model performance is compared across varying the number of TiM and its order. Overall, a single TiM consistently improves the \textit{F1-score} compared to the baseline model without the TiM process. However, this improvement is primarily driven by a consistent increase in \textit{HitRate} across all TiM configurations, although it is accompanied by a degradation in \textit{TrueAlarm}. This trend shows that the TiM process tends to relatively over-predict flood hazards.

Furthermore, as the number of TiM increases, the performance gains diminish. This trend indicates that errors in generated modalities may accumulate and propagate through the chained generation process, negatively affecting subsequent predictions. In addition, the order of modality generation plays a significant role in model performance. For example, the generation sequences involving S2L2A and DEM show noticeable differences in \textit{F1-score} depending on their order. Similarly, combinations of S2L2A and LULC exhibit substantial variation in both \textit{HitRate} and \textit{TrueAlarm}. These findings highlight that not only the choice of modalities but also their order is critical for achieving optimal performance in the TiM framework.

\section{Conclusion and Future Work}

This paper presents ZeroFlood, flood hazard mapping using a single data modality and GeoFMs. Experimental results show that the TerraMind model achieves the best performance among the evaluated models, and its performance can be further improved with TiM settings. Furthermore, our investigation of different TiM configurations suggests that modality generation strategies can extend model performance, but carefully consider the number and order. Future work will evaluate the generalization capability of the model across unseen geographic regions. In addition, incorporating minimal hydrological priors and river-network information into the network may further improve model performance. Collectively, these directions outline a path toward a more general and scalable neural network for flood hazard analysis.


\begin{figure*}[t!]
    \centering
    \includegraphics[width=0.91\textwidth]{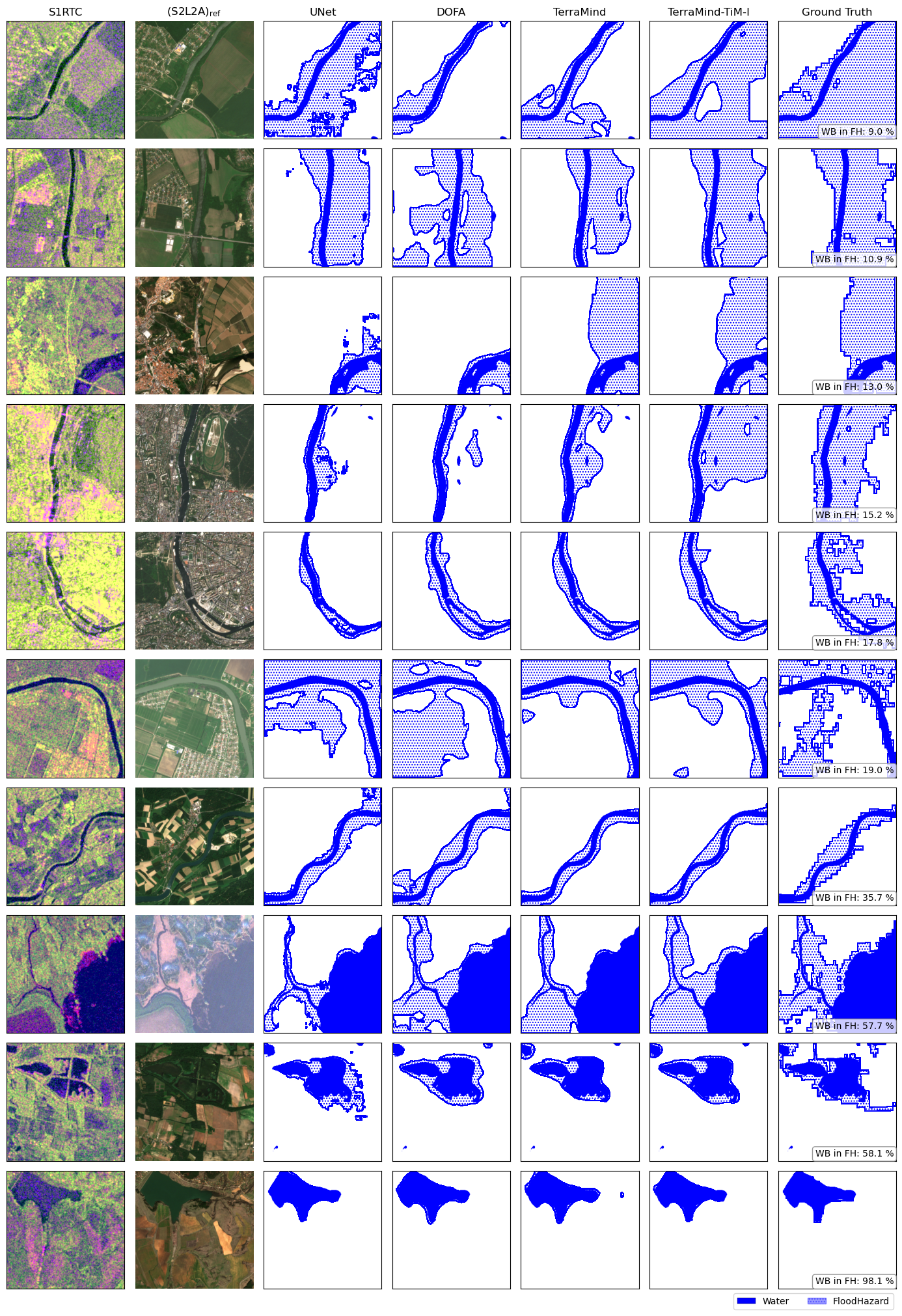}
    \caption{Examples of flood hazard mapping. S1RTC is used as the input modality and is visualized alongside a reference S2RGB image, followed by the predictions of four different models and the ground truth. The trained models predict only flood hazard masks. Water masks are derived from LULC data and are visualized together with the flood hazard masks. A total of 10 samples are selected with different ratios of water bodies within the flood hazard mask (WB in FH), which is shown in the lower-right corner of each ground truth mask.}
  \label{fig:sample_all}
\end{figure*}

\begin{figure*}[t!]
    \centering
    \includegraphics[width=0.65\textwidth]{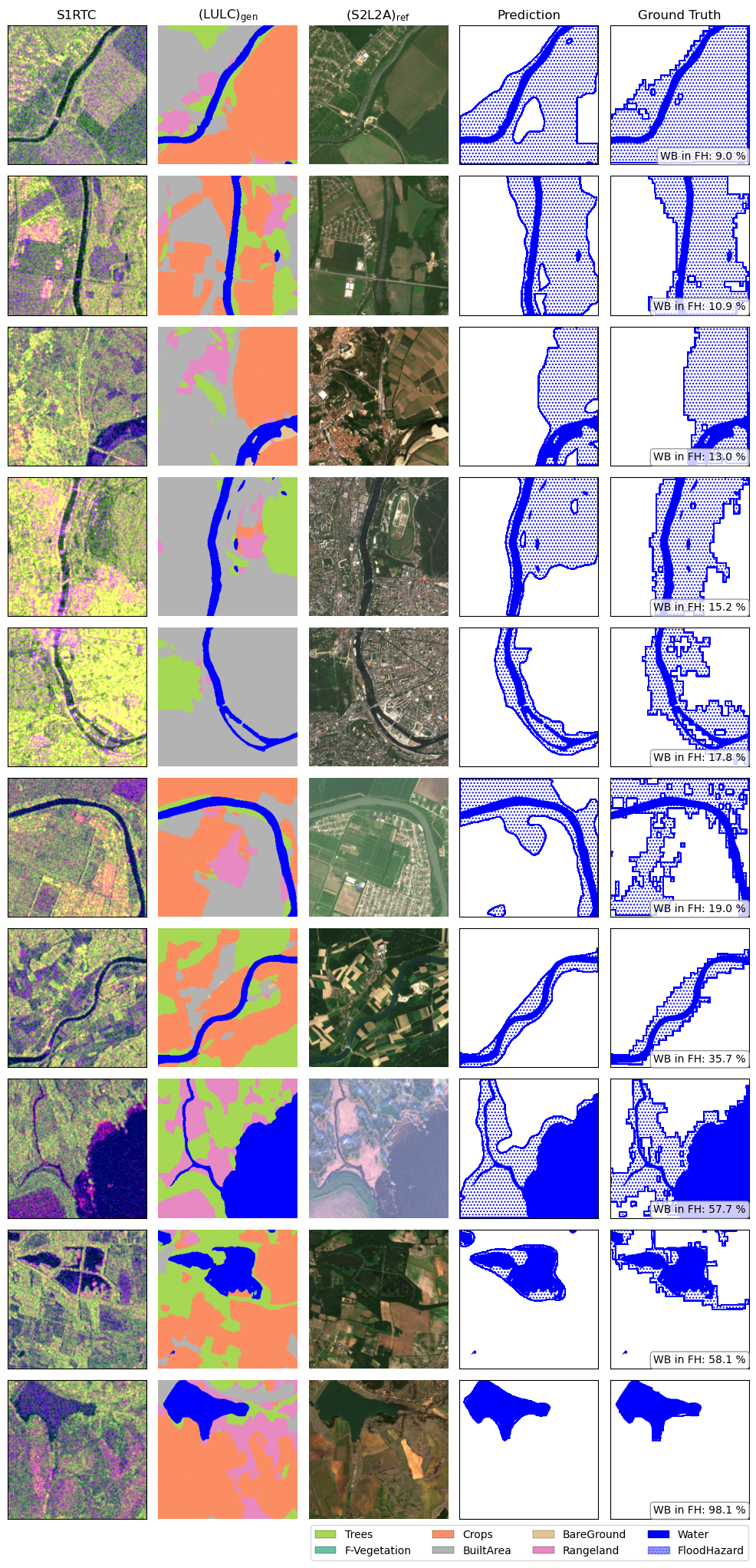}
    \caption{Examples of flood hazard mapping using the TerraMind model with TiM. S1RTC is used as the unimodal input. LULC are TiM-generated modalities and virtually visualized. In addition, the reference S2RGB image, the model predictions, and the ground truth are displayed alongside. The trained model predicts only flood hazard masks. But, water masks are derived from LULC data and are visualized together with the flood hazard masks. A total of 10 samples are selected with different ratios of water bodies within the flood hazard mask (WB in FH), which is shown in the lower-right corner of each ground truth mask.}
    \label{fig:sample_tim}
\end{figure*}

\end{document}

%% file: tabs/dataset_stats.tex
\begin{table}[t!]
\centering
\fontsize{10}{12}\selectfont
\begin{tabularx}{\columnwidth}{cccccc}
\hline
\textit{Split} & \textit{Samples} & \multicolumn{4}{c}{\textit{WB in FH (\%)}} \\
      &         & \textit{Median} & \textit{Mean} & \textit{Min} & \textit{Max} \\
\hline
Train & 20868 & 87.31 & 75.48 & 4.87 & 100 \\
Val   & 5217  & 86.88 & 75.17 & 5.07 & 100 \\
Test  & 226   & 90.06 & 78.61 & 6.61 & 99.92 \\
\hline
\end{tabularx}
\caption{Dataset statistics. \textit{WB in FH} denotes the percentage of water bodies located within flood hazard masks. It is close to 100 when the flood hazard mask is identical to the water body.}
\label{tab:dataset_stats}
\end{table}

%% file: tabs/main_metrics.tex
\begin{table}[t!]
\centering
\fontsize{10}{12}\selectfont
\begin{tabularx}{\columnwidth}{ccccc}
\hline
\textit{Model}         & \textit{F1-score}             & \textit{HitRate}             & \textit{TrueAlarm}            \\ \hline
UNet          & 84.66          & 81.00          & 88.67          \\
ViT           & 84.64          & 81.46          & 88.08          \\ \hline
SSL4EO-MAE    & 85.23          & 81.39          & 89.44          \\
CROMA         & 83.63          & 79.72          & 87.95          \\
DOFA          & 86.90          & 84.80          & 89.11          \\ \hline
TerraMind     & {\ul 88.36}    & {\ul 86.77}    & \textbf{90.02}    \\
TerraMind-TiM-l & \textbf{89.12} & \textbf{88.45} & {\ul 89.80} \\ \hline
\end{tabularx}
\caption{Performance comparison between models. The best score in each column is shown in bold, and the second-best score is underlined.}
\label{tab:main_metrics}
\end{table}

%% file: EUSAR-2026-Submission-v2.bbl
\begin{thebibliography}{9}\leftskip1mm\advance\labelsep\leftskip
\bibitem{A19} Aronsson-Storrier M. UN office for disaster risk reduction (2019). YB Int'l Disaster L. Online. 2021;2:377.
\bibitem{ASC20} Albano R, Samela C, Crăciun I, Manfreda S, Adamowski J, Sole A, Sivertun Å, Ozunu A. Large scale flood risk mapping in data scarce environments: An application for Romania. Water. 2020 Jun 26;12(6):1834.
\bibitem{BFM25} Blumenstiel B, Fraccaro P, Marsocci V, Jakubik J, Maurogiovanni S, Czerkawski M, Sedona R, Cavallaro G, Brunschwiler T, Moreno JB, Longépé N. Terramesh: A planetary mosaic of multimodal earth observation data. InProceedings of the Computer Vision and Pattern Recognition Conference 2025 (pp. 2394-2402).
\bibitem{BIP20} Baky MA, Islam M, Paul S. Flood hazard, vulnerability and risk assessment for different land use classes using a flow model. Earth Systems and Environment. 2020 Mar;4(1):225-44.
\bibitem{DAB21} Dottori F, Alfieri L, Bianchi A, Skoien J, Salamon P. A new dataset of river flood hazard maps for Europe and the Mediterranean Basin region. Earth System Science Data Discussions. 2021 Jan 12;2021:1-35.
\bibitem{DBK20} Dosovitskiy A, Beyer L, Kolesnikov A, Weissenborn D, Zhai X, Unterthiner T, Dehghani M, Minderer M, Heigold G, Gelly S, Uszkoreit J. An image is worth 16x16 words: Transformers for image recognition at scale. arXiv preprint arXiv:2010.11929. 2020 Oct 22.
\bibitem{DFC24} Dionelis N, Fibaek C, Camilleri L, Luyts A, Bosmans J, Saux BL. Evaluating and benchmarking foundation models for earth observation and geospatial ai. arXiv preprint arXiv:2406.18295. 2024 Jun 26.
\bibitem{DYF21}Dong S, Yu T, Farahmand H, Mostafavi A. A hybrid deep learning model for predictive flood warning and situation awareness using channel network sensors data. Computer‐Aided Civil and Infrastructure Engineering. 2021 Apr;36(4):402-20.
\bibitem{FEA24} Fereshtehpour M, Esmaeilzadeh M, Alipour RS, Burian SJ. Impacts of DEM type and resolution on deep learning-based flood inundation mapping. Earth Science Informatics. 2024 Apr;17(2):1125-45.
\bibitem{FMG23} Fuller A, Millard K, Green J. CROMA: Remote sensing representations with contrastive radar-optical masked autoencoders. Advances in Neural Information Processing Systems. 2023 Dec 15;36:5506-38.
\bibitem{HG21}Hashemi-Beni L, Gebrehiwot AA. Flood extent mapping: An integrated method using deep learning and region growing using UAV optical data. IEEE Journal of Selected Topics in Applied Earth Observations and Remote Sensing. 2021 Jan 14;14:2127-35.
\bibitem{JYB25} Jakubik J, Yang F, Blumenstiel B, Scheurer E, Sedona R, Maurogiovanni S, Bosmans J, Dionelis N, Marsocci V, Kopp N, Ramachandran R. Terramind: Large-scale generative multimodality for earth observation. arXiv preprint arXiv:2504.11171. 2025 Apr 15.
\bibitem{LLW23} Li W, Lee H, Wang S, Hsu CY, Arundel ST. Assessment of a new GeoAI foundation model for flood inundation mapping. InProceedings of the 6th ACM SIGSPATIAL International workshop on AI for geographic knowledge discovery 2023 Nov 13 (pp. 102-109).
\bibitem{LXC16} Li L, Xu T, Chen Y. Improved urban flooding mapping from remote sensing images using generalized regression neural network-based super-resolution algorithm. Remote Sensing. 2016 Jul 28;8(8):625.
\bibitem{MJB25} Marsocci V, Jia Y, Bellier GL, Kerekes D, Zeng L, Hafner S, Gerard S, Brune E, Yadav R, Shibli A, Fang H. Pangaea: A global and inclusive benchmark for geospatial foundation models. arXiv preprint arXiv:2412.04204. 2024 Dec 5.
\bibitem{MRM23} Mehravar S, Razavi-Termeh SV, Moghimi A, Ranjgar B, Foroughnia F, Amani M. Flood susceptibility mapping using multi-temporal SAR imagery and novel integration of nature-inspired algorithms into support vector regression. Journal of Hydrology. 2023 Feb 1;617:129100.
\bibitem{NC20} Nones M, Caviedes‐Voullième D. Computational advances and innovations in flood risk mapping. Journal of Flood Risk Management. 2020 Dec 1;13(4).
\bibitem{PLD22} Prasad P, Loveson VJ, Das B, Kotha M. Novel ensemble machine learning models in flood susceptibility mapping. Geocarto International. 2022 Aug 18;37(16):4571-93.
\bibitem{RFB15} Ronneberger O, Fischer P, Brox T. U-net: Convolutional networks for biomedical image segmentation. InInternational Conference on Medical image computing and computer-assisted intervention 2015 Oct 5 (pp. 234-241). Cham: Springer international publishing.
\bibitem{RS20} Rentschler J, Salhab M. People in Harm’s Way. Flood Exposure and Poverty in. 2020 Oct;189.
\bibitem{SJM21} Sidek LM, Jaafar AS, Majid WH, Basri H, Marufuzzaman M, Fared MM, Moon WC. High-resolution hydrological-hydraulic modeling of urban floods using InfoWorks ICM. Sustainability. 2021 Sep 14;13(18):10259.
\bibitem{SKH23} Seydi ST, Kanani-Sadat Y, Hasanlou M, Sahraei R, Chanussot J, Amani M. Comparison of machine learning algorithms for flood susceptibility mapping. Remote Sensing. 2022 Dec 29;15(1):192.
\bibitem{SSB15}Sampson CC, Smith AM, Bates PD, Neal JC, Alfieri L, Freer JE. A high‐resolution global flood hazard model. Water resources research. 2015 Sep;51(9):7358-81.
\bibitem{TSK21} Tellman B, Sullivan JA, Kuhn C, Kettner AJ, Doyle CS, Brakenridge GR, Erickson TA, Slayback DA. Satellite imaging reveals increased proportion of population exposed to floods. Nature. 2021 Aug 5;596(7870):80-6.
\bibitem{WBX23} Wang Y, Braham NA, Xiong Z, Liu C, Albrecht CM, Zhu XX. SSL4EO-S12: A large-scale multimodal, multitemporal dataset for self-supervised learning in Earth observation [Software and Data Sets]. IEEE Geoscience and Remote Sensing Magazine. 2023 Sep 25;11(3):98-106.
\bibitem{XLZ18} Xiao T, Liu Y, Zhou B, Jiang Y, Sun J. Unified perceptual parsing for scene understanding. InProceedings of the European conference on computer vision (ECCV) 2018 (pp. 418-434).
\bibitem{XWZ25} Xiong Z, Wang Y, Zhang F, Stewart AJ, Hanna J, Borth D, Papoutsis I, Le Saux B, Camps-Valls G, Zhu XX. Neural plasticity-inspired foundation model for observing the earth crossing modalities. CoRR. 2024 Jan 1.
\bibitem{ZSH16} Zarzar C, Siddique R, Hosseiny H, Gomez M. Quantifying uncertainty in flood inundation mapping using streamflow ensembles and hydraulic modeling techniques. Natl. WATER Cent. Innov. Progr. SUMMER Inst. Rep. 2016 Oct 20;4:71.



\end{thebibliography}
